\pdfoutput=1
\documentclass[11pt]{article}

\usepackage{naacl2021}

\usepackage{times}
\usepackage{latexsym}

\usepackage[T1]{fontenc}
\usepackage[utf8]{inputenc}

\usepackage{microtype}

\usepackage{fixltx2e}

\usepackage{graphicx}
\usepackage{subcaption}

\usepackage{booktabs}
\usepackage{multirow}
\usepackage{siunitx}
\usepackage{adjustbox}

\usepackage{xurl}

\title{Cost-effective Deployment of BERT Models in a Serverless Environment}

\author{
  Katar\'{i}na Benešová \thanks{~~Equal contribution} \\
  Slido \\
  \texttt{kbenesova@slido.com} \\\And
  Andrej Švec \footnotemark[1] \\
  Slido \\
  \texttt{asvec@slido.com} \\\And
  Marek Šuppa \footnotemark[1] \\
  Slido \\
  \texttt{msuppa@slido.com}}

\begin{document}
\maketitle
\begin{abstract}
In this study we demonstrate the viability of deploying BERT-style models to serverless environments in a production setting. Since the freely available pre-trained models are too large to be deployed in this way, we utilize knowledge distillation and fine-tune the models on proprietary datasets for two real-world tasks: sentiment analysis and semantic textual similarity. As a result, we obtain models that are tuned for a specific domain and deployable in serverless environments. The subsequent performance analysis shows that this solution results in latency levels acceptable for production use and that it is also a cost-effective approach for small-to-medium size deployments of BERT models, all without any infrastructure overhead.
\end{abstract}

\section{Introduction}

Machine learning models are notoriously hard to bring to production environments. One of the reasons behind is the large upfront infrastructure investment it usually requires. This is particularly the case with large pre-trained language models, such as BERT \cite{devlin2018bert} or GPT \cite{radford2019language} whose size requirements make them difficult to deploy even when infrastructure investment is not of concern.

At the same time, the serverless architecture with minimal maintenance requirements, automatic scaling and attractive cost, is becoming more and more popular in the industry. It is very well suited for stateless applications such as model predictions, especially in cases when the prediction load is unevenly distributed. Since the serverless platforms have strict limits, especially on the size of the deployment package, it is not immediately obvious it may be a viable platform for deployment of models based on large pre-trained language models.

In this paper we describe our experience with deploying BERT-based models to serverless environments in a production setting. We consider two tasks: sentiment analysis and semantic textual similarity. While the standard approach would be to fine-tune the pre-trained models, this would not be possible in our case, as the resulting models would be too large to fit within the limits imposed by serverless environments. Instead, we adopt a knowledge distillation approach in combination with smaller BERT-based models. We show that for some of the tasks we are able to train models that are an order of magnitude smaller while reporting performance similar to that of the larger ones. 

Finally, we also evaluate the performance of the deployed models. Our experiments show that their latency is acceptable for production environments. Furthermore, the reported costs suggest it is a very cost-effective option, especially when the expected traffic is small-to-medium in size (a few requests per second) and potentially unevenly distributed.

\section{Related work}
Despite a number of significant advances in various NLP approaches over the recent years, one of the limiting factors hampering their adoption is the large number of parameters that these models have, which leads to large model size and increased inference time. This may limit their use in resource-constrained mobile devices or any other environment in which model size and inference time is the limiting factor, while negatively affecting the environmental costs of their use \cite{strubell2019energy} .

This has led to a significant body of work focusing on lowering both the model size and inference time, while incurring minimal performance penalty. One of the most prominent approaches include Knowledge Distillation \cite{bucilu2006model, hinton2015distilling}, in which a smaller model (the ''student'') is trained to reproduce the behavior of a larger model (the ''teacher''). It was used to produce smaller BERT alternatives, such as:

\begin{itemize}
    %\item \textbf{DistilBERT} \cite{sanh2019distilbert}, which reduces the number of layers by 2,  applies knowledge distillation on the last layer and finetunes the resulting model on downstream tasks,
    \item \textbf{TinyBERT} \cite{jiao2019tinybert}, which appropriates the knowledge transfer method to the Transformer architecture and applies it in both the pretraining and downstream fine-tuning stage. The resulting model is more than 7x smaller and 9x faster in terms of inference.
    \item \textbf{MobileBERT} \cite{sun2020mobilebert}, which only uses knowledge distilation in the pre-training stage and reduces the model's width (layer size) as opposed to decreasing the number of layers it consists of. The final task-agnostic model is more than 3x smaller and 5x faster than the original BERT\textsubscript{BASE}. 
\end{itemize}
 
When decreasing the model size leads to decreased latency, it can also have direct business impact. This has been demonstrated by Google, which found out that increasing web search latency from 100~ms to 400~ms reduced the number of searches per user by 0.2~\% to 0.6~\% \cite{brutlag2009speed}. A similar experiment done by Booking.com has shown that an increase in latency of about 30~\% results in about 0.5 percentage points decrease in conversion rates, which the authors report as a \emph{''relevant cost for our business''} \cite{bernardi2019150}.

Each serverless platform has its specifics, which can have different impact on different use cases. Various works, such as \cite{back2018using, wang2018peeking, lee2018evaluation}, provide a comparison of performance differences between the available platforms. In order to evaluate specific use cases, various benchmark suites have been introduced such as FunctionBench \cite{kim2019functionbench}, which includes language generation as well as sentiment analysis test case.

Possibly the closest published work comparable to ours is \cite{tu2018pay}, in which the authors demonstrate the deployment of neural network models, trained for short text classification and similarity tasks in a serverless context. Since at the time of its publication the PyTorch deployment ecosystem has been in its nascent stages, the authors had to build it from source, which complicates practical deployment.

To the best of our knowledge, our work is the first to show the viability of deploying large pre-trained language models (such as BERT and its derivatives) in the serverless environment.

\section{Serverless environments}
\label{sec:serverless}

Serverless environments offer a convenient and affordable way of deploying a small piece of code. A survey by O’Reilly Media \cite{oreilly2019serverless} shows that the adoption of serverless was successful for the majority of the respondents' companies. They recognize reduced operational costs, automatic scaling with demand and elimination of concerns for server maintenance as the main benefits.

Since the functions deployed in a serverless environment share underlying hardware, OS and runtime \cite{lynn2017preliminary}, there are naturally numerous limitations to what can be run in such environment.
The most pronounced ones include:

\begin{itemize}
    \item \textbf{Maximum function size}, mostly limited to a few hundreds of MBs (although some providers do not have this limitation). In the context of deployment of a machine learning model, this can significantly limit the model size as well as the selection of libraries to be used to execute the model.
    % \item \textbf{Maximum execution time}, usually on the order of tens of minutes, making serverless functions unsuitable for longer jobs.
    \item \textbf{Maximum memory} of a few GBs slows down or makes it impossible to run larger models. 
    \item \textbf{No acceleration.} Serverless environments do not support GPU or TPU acceleration which can significantly increase the inference time for larger models.
    %\item \textbf{Request size} of a few MBs could pose a problem for some model inputs, such as images. However, it should not be limiting for textual inputs.
\end{itemize}

A more detailed list of the main limitations of the three most common serverless providers can be found in Table~\ref{tab:serverless-providers}. It suggests that any model deployed in this environment will need to be small in size and have minimal memory requirements. These requirements significantly limit the choice of models appropriate for this environment and warrants a specific training regimen, which we describe in the next section.

\begin{table}
\centering
\begin{tabular}{lccc}
\hline
& \textbf{AWS} &  \textbf{Azure} & \textbf{GCP} \\
\hline
Function size  & 250MB\footnotemark & - & 500MB  \\
Execution time & 15min  & - & 9min \\
Memory         & 10GB & 14GB &  8GB \\
Request size   & 6MB  & 100MB & 10MB \\
\hline
\end{tabular}
\caption{Limitations of the three main serverless providers: Amazon Web Services (AWS), Microsoft Azure (Azure) and Google Cloud Platform (GCP).}
\label{tab:serverless-providers}
\end{table}

\footnotetext{Recently, a new way of deployment was added, allowing to deploy a container of size up to 10~GB.}

\begin{figure*}
\centering
\includegraphics[width=0.8\textwidth]{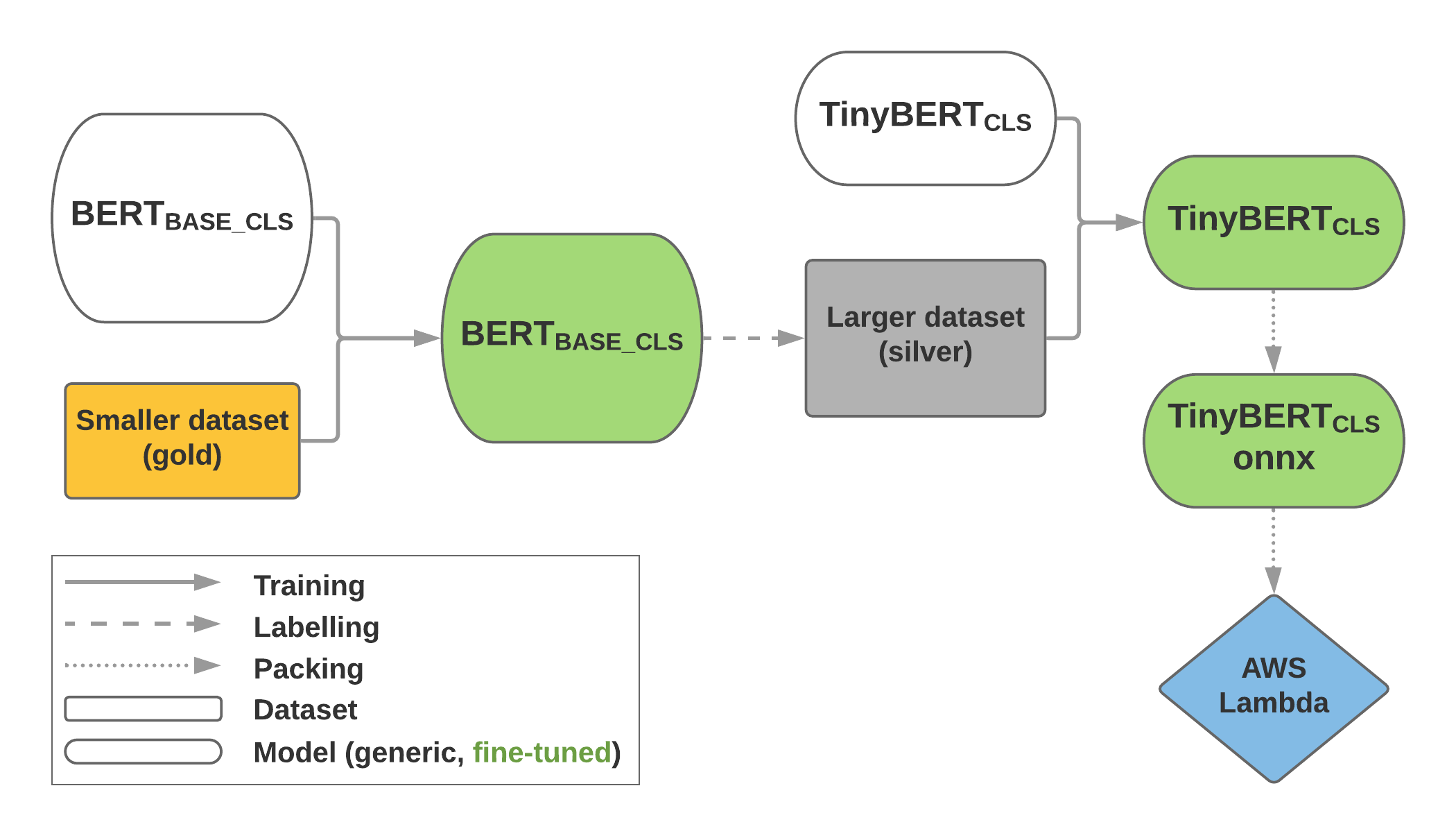}
\caption{Schema of the distillation pipeline of BERT\textsubscript{BASE} for sentiment analysis. BERT\textsubscript{BASE\_CLS} is fine-tuned on the gold dataset and then used for labelling a large amount of data (silver dataset) that serves as a training set for distillation to TinyBERT. The distilled model is exported to the ONNX format and deployed to AWS Lambda (see Section~\ref{deployment}). The same pipeline was executed for MobileBERT.}
\label{fig:pipeline-sentiment}
\end{figure*}

\section{Model training}

In the two case studies presented in this section, we first consider BERT-provided classification token (\texttt{[CLS]} token) an aggregate representation of a short text (up to 300 characters) for the sentiment analysis task. Secondly, we utilize the embeddings produced by Sentence-BERT (SBERT) \cite{reimers2019sentence} for estimating the semantic similarity of a pair of short texts.

Since deploying even the smaller BERT\textsubscript{BASE} with over 400MB in size is not possible in our setup, in the following cases studies we explore several alternative approaches, such as knowledge distillation into smaller models or training a smaller model directly. To do so, we use TinyBERT \cite{jiao2019tinybert} and MobileBERT \cite{sun2020mobilebert} having about 56~MB and 98~MB in size, respectively.

\subsection{BERT for sentiment analysis} \label{bert_sentiment}
One of the direct applications of the special \texttt{[CLS]} token of BERT is the analysis of sentiment \cite{li2019exploiting}. We formulate this problem as classification into three categories: \textit{Positive}, \textit{Negative} and \textit{Neutral}.

The task is divided into two stages: first, we fine-tune BERT\textsubscript{BASE} using a labelled domain-specific dataset of 68K training examples and 9K examples for validation. Then we proceed with knowledge distillation into a smaller model with faster inference: we label a large amount of data by the fine-tuned BERT\textsubscript{BASE} and use the dataset to train a smaller model with a BERT-like architecture. The distillation pipeline is illustrated in Figure~\ref{fig:pipeline-sentiment}.

\subsubsection{Fine-tuning BERT\textsubscript{BASE}} \label{bert_finetuning}
To utilize BERT\textsubscript{BASE} for a classification task, an additional head must be added on top of the Transformer blocks, i.e. a linear layer on top of the pooled output. The additional layer typically receives only the representation of the special \texttt{[CLS]} token as its input. To obtain the final prediction, the output of this layer is passed through a Softmax layer producing the probability distribution over the predicted classes.

We fine-tuned BERT\textsubscript{BASE} for sequence classification (BERT\textsubscript{BASE\_CLS}) with this adjusted architecture for our task using a labelled dataset of size 68K consisting of domain-specific data. We trained the model for 8 epochs using AdamW optimizer with small learning rate $3\times10^{-5}$, L2 weight decay of 0.01 and batch size 128.

To cope with the significant class imbalance\footnote{About 82\% of the dataset were \textit{Neutral} examples, 10\% \textit{Negative} and 8\% \textit{Positive}.} and to speed up the training, we sampled class-balanced batches in an under-sampling fashion, while putting the examples of similar length together (for the sake of a more effective processing of similarly padded data). Using this method, we were able to at least partially avoid over-fitting on the largest class and reduce the training time about 2.5 times.

We also tried an alternative fine-tuning approach by freezing BERT\textsubscript{BASE} layers and attaching a small trainable network on top of it. For the trainable part, we experimented with 1-layer bidirectional GRU of size 128 with dropout of 0.25 plus a linear layer and Softmax output. BERT\textsubscript{BASE\_CLS} outperformed this approach significantly. 

The accuracy evaluation of both fine-tuned BERT\textsubscript{BASE} models on the validation dataset can be found in Table~\ref{tab:sentiment-eval}. In order to meet the function size requirements of the target serverless environments, we proceed to the knowledge distillation stage.

\subsubsection{Knowledge distillation to smaller BERT models}
Having access to virtually unlimited supply of unlabelled domain-specific examples, we labelled almost 900K of them by the fine-tuned BERT\textsubscript{BASE\_CLS} "teacher" model and used them as ground truth labels for training a smaller "student" model. We experimented with MobileBERT and even smaller TinyBERT as the student models since these are, in comparison to BERT\textsubscript{BASE}, 3 and 7 times smaller in size, respectively.

During training, we sampled the batches in the same way as in Section \ref{bert_finetuning}, except for a smaller batch size of 64. We trained the model for a small number of epochs using AdamW optimizer with learning rate $2\times10^{-5}$, weight decay 0.01 and early stopping after 3 epochs in case of TinyBERT and one epoch for MobileBERT (in the following epochs the models no longer improved on the validation set). 

For evaluation we used the same validation dataset as for the fine-tuned BERT\textsubscript{BASE\_CLS} described in \ref{bert_sentiment}. The performance comparison is summarized in Table~\ref{tab:sentiment-eval}. We managed to distill the model knowledge into the significantly smaller TinyBERT with only 0.02 points decrease in F1 score (macro-averaged). In case of MobileBERT we were able to match the performance of BERT\textsubscript{BASE\_CLS}. These results suggest that the large language models might not be necessary for classification tasks in a real-life scenario. 

\begin{table}
\centering
\begin{tabular}{lrcc}
\hline
\textbf{Model} & \textbf{Size} (MB) & \textbf{F1} \\
\hline
BERT\textsubscript{BASE} + GRU               & 426 & 0.75 \\
BERT\textsubscript{BASE\_CLS}  & 420 & \textbf{0.84} \\
TinyBERT \color{gray}{\footnotesize
 (distilled)}                & 56  & 0.82 \\
MobileBERT \color{gray}{\footnotesize
 (distilled)}              & 98 & \textbf{0.84} \\
\hline
\end{tabular}
\caption{Comparison of fine-tuned BERT models and smaller distilled models on the validation dataset (macro-averaged F1 score). The slight decrease in TinyBERT's performance is an acceptable trade-off for the significant size reduction.}
\label{tab:sentiment-eval}
\end{table}

\subsection{Sentence-BERT for semantic textual similarity}

The goal of our second case study was to train a model that would generate dense vectors usable for semantic textual similarity (STS) task in our specific domain and be small enough to be deployed in a serverless environment. The generated vectors would then be indexed and queried as part of a duplicate text detection feature of a real-world web application. To facilitate this use-case, we use Sentence-BERT (SBERT) \cite{reimers2019sentence}.

While the SBERT architecture currently reports state-of-the-art performance on the sentence similarity task, all publicly available pre-trained SBERT models are too large for serverless deployment. The smallest one available is SDistilBERT\textsubscript{BASE} with on-disk size of 255~MB. We therefore had to train our own SBERT model based on smaller BERT alternatives. We created the smaller SBERT models by employing the TinyBERT and MobileBERT into the SBERT architecture, i.e. by adding an embedding averaging layer on top of the BERT model.

\begin{figure*}[tb]
\centering
\includegraphics[width=0.85\textwidth]{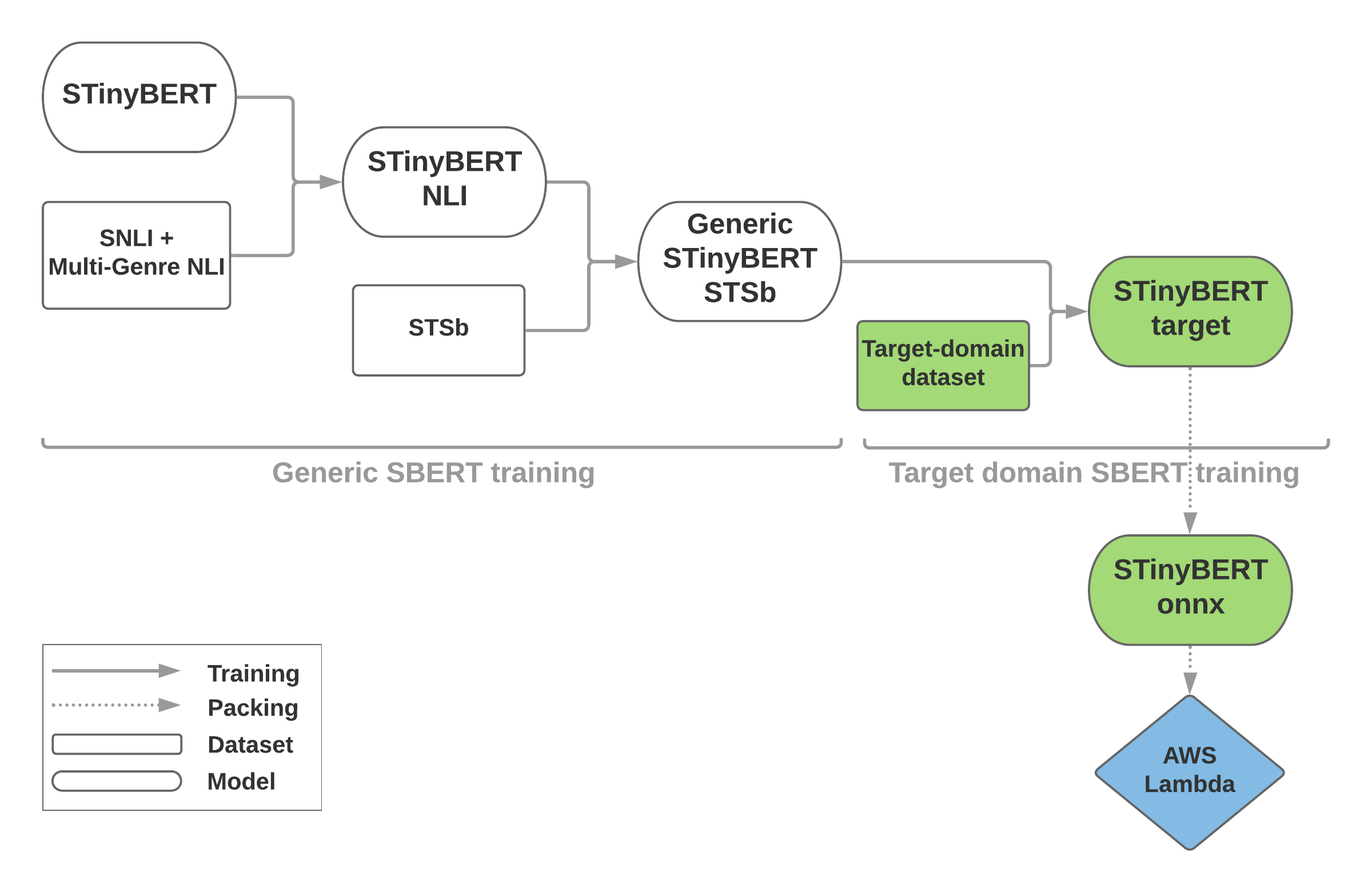}
\caption{Schema of the fine-tuning pipeline of STinyBERT for STS task. In the first stage, STinyBERT is fine-tuned on NLI and STSb datasets to obtain Generic STinyBERT. In the second phase, the model is trained further on the target-domain dataset, exported to the ONNX format and deployed to AWS Lambda (see Section~\ref{deployment}). The same pipeline was executed for SMobileBERT. SBERT\textsubscript{BASE} was only fine-tuned on target domain dataset.}
\label{fig:pipeline-similarity}
\end{figure*}

In order to make the smaller SBERT models perform on the STS task, we fine-tune them in two stages. Firstly, we fine-tune them on standard datasets to obtain a smaller version of the generic SBERT model and then we fine-tune them further on the target domain data. The fine-tuning pipeline is visualized in Figure \ref{fig:pipeline-similarity}.

\subsubsection{Generic SBERT fine-tuning}
To obtain a smaller version of SBERT, we followed the the SBERT training method as outlined in \cite{reimers2019sentence}. We first fine-tuned a smaller SBERT alternative on a combination of SNLI \cite{bowman2015snli} (dataset of sentence pairs labeled for entailment, contradiction, and semantic independence) and Multi-Genre NLI \cite{williams2018multinli} (dataset of both written and spoken speech in a wide range of styles, degrees of formality, and topics) datasets. 

We observed the best results when fine-tuning the model for 4 epochs with early stopping based on validation set performance, batch size 16, using Adam optimizer with learning rate $2 \times 10^{-5}$ and a linear learning rate warm-up over 10~\% of the total training batches.

Next, we continued fine-tuning the model on the STSbenchark (STSb) dataset \cite{cer2017stsb} using the same approach, except for early stopping based on STSb development set performance and a batch size of 128.

\subsubsection{Target domain fine-tuning}

Once we obtained a small enough generic SBERT model, we proceeded to fine-tune it on examples from the target domain. We experimented with two approaches: fine-tuning the model on a small gold dataset and generating a larger silver dataset.

\paragraph{Dataset.}
We worked with a balanced training set of 2856 pairs. Each pair was assigned to one of three classes: \textit{duplicate} (target cosine similarity 1), \textit{related} (0.5) or \textit{unrelated} (0). The classes were assigned semi-automatically. \textit{Duplicate} pairs were created by back-translation \cite{sennrich2016backtranslation} using the translation models released as part of the OPUS-MT project \cite{tiedemann:helsinki}. \textit{Related} pairs were pre-selected and expertly annotated and \textit{unrelated} pairs were formed by pairing random texts together.

Validation and test sets were composed of 665 and 696 expertly annotated pairs, respectively. These sets were not balanced due to the fact that finding \textit{duplicate} pairs manually is far more difficult than finding \textit{related} or \textit{unrelated} pairs, which stems from the nature of the problem. That is why \textit{duplicate} class forms only approximately 13~\% of the dataset, whereas \textit{related} and \textit{unrelated} classes each represent roughly 43~\%.

% As it is often the case with production systems, apart from small sets of annotated data, we also had access to a lot of data without annotations.

\paragraph{Fine-tuning on plain dataset.}
We first experimented with fine-tuning the generic SBERT model on the train set of the target domain dataset. We call the output model SBERT target. We fine-tuned it for 8 epochs with early stopping based on validation set performance, batch size 64, Adam optimizer with learning rate $2 \times 10^{-5}$ and a linear learning rate warm-up over 10~\% of the total training batches.

\paragraph{Extending the dataset.}
Since we had a lot of data without annotations available, we also experimented with extending the dataset and fine-tuning Augmented SBERT \cite{thakur2020augsbert}.

We pre-selected 379K duplicate candidates using BM25 \cite{Amati2009} and annotated them using a pre-trained cross-encoder based on RoBERTa\textsubscript{LARGE}. In the annotated data, low similarity values were majorly prevalent (median similarity was 0.18). For this reason, we needed to balance the dataset by undersampling the similarity bins with higher number of samples to get to a final balanced dataset of 32K pairs. We refer to the original expert annotations as gold data and to the cross-encoder annotations as silver data.

% \begin{figure}
% \centering
% \includegraphics[width=\columnwidth]{silver-sim-hist.png}
% \caption{Histogram of similarities of silver dataset pairs predicted by the RoBERTa\textsubscript{LARGE} based cross-encoder.}
% \label{fig:silver-sim-hist}
% \end{figure}

After creating the silver dataset, we first fine-tuned the model on the silver data and then on the gold data. We call the model fine-tuned on augmented target dataset AugSBERT. Correct hyperparameter selection was crucial for a successful fine-tuning. It was especially necessary to lower the learning rate for the final fine-tuning on the gold data and set the right batch sizes. For the silver dataset we used a learning rate of $2 \times 10^{-5}$ and batch size of 64. For the final fine-tuning on the gold dataset we used a lower learning rate of $2 \times 10^{-6}$ and a batch size of 16.

\subsubsection{Results}

%We report the test set performances of different models obtained after individual fine-tuning stages in Table~\ref{tab:sbert-results}.

As we can see in Table~\ref{tab:sbert-results}, smaller BERT alternatives can compete with SBERT\textsubscript{BASE}. AugSMobileBERT manages to reach 93~\% of the performance of SBERT\textsubscript{BASE} on the target dataset while being more than 3 times smaller in size.

We believe that the lower performance of smaller models is not only caused by the them having less parameters, but it also essentially depends on the size of the model's output dense vector. TinyBERT's output embedding size is 312 and MobileBert's is 512, whereas BERT\textsubscript{BASE} outputs embeddings of size 768. This would in line with the findings published in \cite{wieting2019no} which state that even random projection to a higher dimension leads to increased performance.

\begin{table}[tbh]
\centering
\begin{tabular}{lcc}
\hline
\textbf{Model} & \textbf{STSb} & \textbf{Target} \\
\hline
STinyBERT NLI    & 72.86 & 46.29 \\
SMobileBERT NLI  & 78.29 & 52.08 \\
SBERT\textsubscript{BASE} NLI  & 77.03 & 52.44 \\
\hline
STinyBERT STSb   & 76.76 & 53.89 \\
SMobileBERT STSb  & 81.52 & 59.05 \\
SBERT\textsubscript{BASE} STSb  & 85.35 & 65.87 \\
\hline
STinyBERT target      & 75.49 & 53.29 \\
SMobileBERT target    & 79.56 & 59.27 \\
SBERT\textsubscript{BASE} target & 82.52 & 64.20 \\
\hline
AugSTinyBERT target            & 73.88          & 54.34 \\
\textbf{AugSMobileBERT target} & \textbf{80.47} & \textbf{61.75} \\
AugSBERT\textsubscript{BASE} target             & 82.98          & 64.14 \\
\hline
\end{tabular}
\caption{Spearman rank correlation between the cosine similarity of dense vectors and true labels measured for individual models on the test set of the STSbenchmark dataset (STSb column) and on the test set of the target domain dataset (Target column). The values are multiplied by 100 for convenience. We also present SBERT\textsubscript{BASE} performance as baseline. The model with the best performance on the target domain dataset, that is also deployable in serverless environment, is highlighted.}
\label{tab:sbert-results}
\end{table}

\section{Deployment}
\label{deployment}

As described in Section~\ref{sec:serverless}, numerous limitations must be satisfied when deploying a model to a serverless environment, among which the size of the deployment package is usually the major one. The deployment package consists of the function code, runtime libraries and in our case a model.

\begin{figure*}[tb]
    \centering
    \begin{subfigure}{\columnwidth}
    \includegraphics[width=\textwidth]{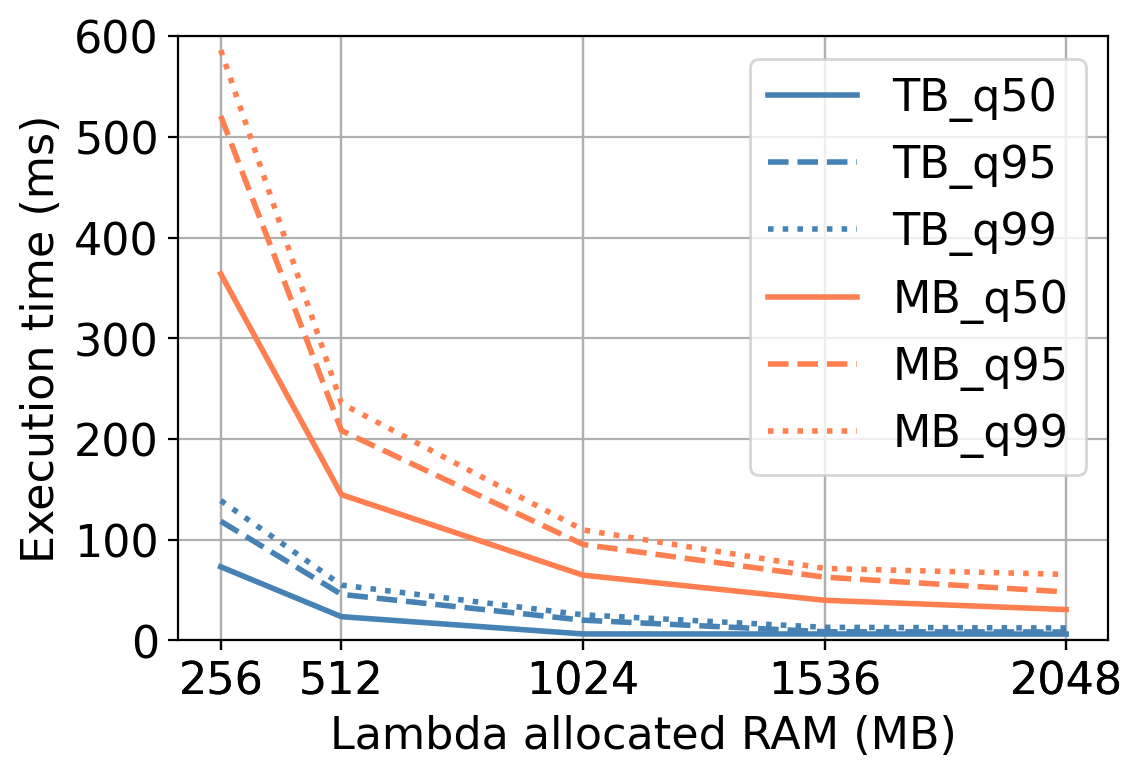}
    \caption{Sentiment analysis.}
    \label{fig:sentiment-duration}
    \end{subfigure}
\quad
    \begin{subfigure}{\columnwidth}
    \includegraphics[width=\textwidth]{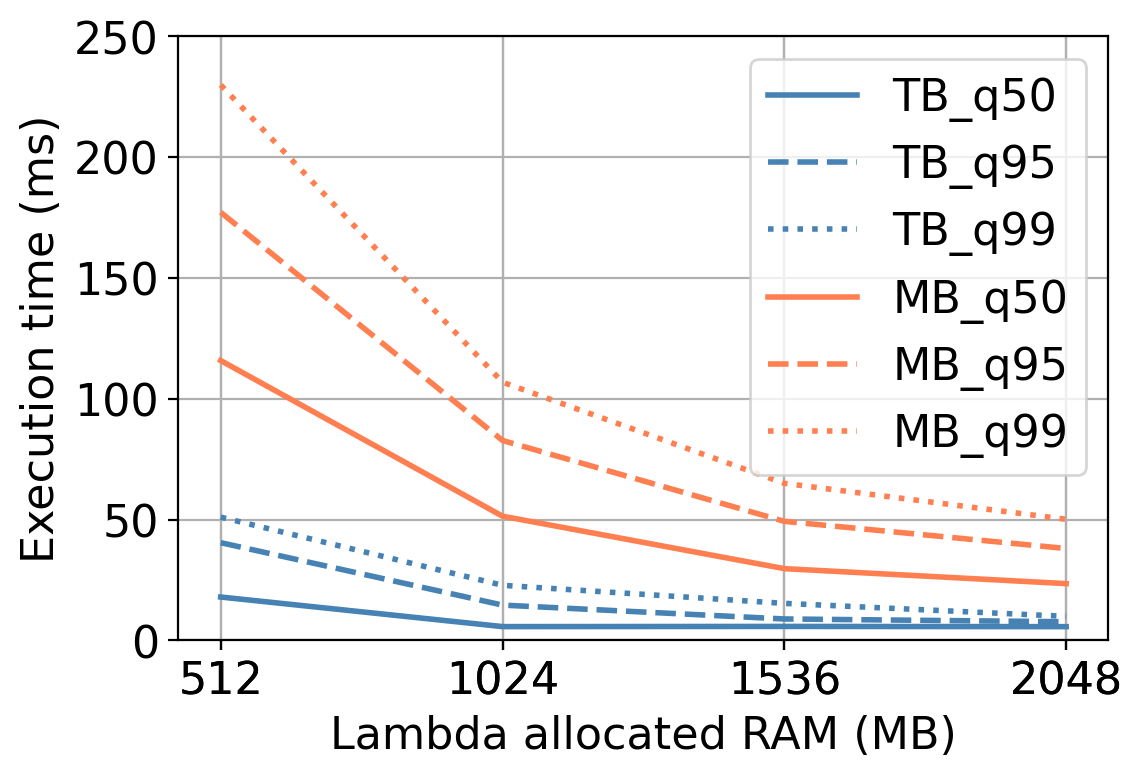}
    \caption{SBERT encoding.}
    \label{fig:sbert-duration}
    \end{subfigure}

    \caption{Results of performance tests of trained models deployed in AWS Lambda. Execution time is denoted in miliseconds (ms). TB stands for TinyBERT, MB for MobileBERT. q50, q95 and q99 denote the 0.5, 0.95 and 0.99 quantiles, respectively.}
    \label{fig:all-duration}
\end{figure*}

\subsection{Model inference engine}
In order to fit all of the above in a few hundreds of MBs allowed in the serverless environments, standard deep learning libraries cannot be used: the standard PyTorch wheel has 400~MB \cite{paszke2019pytorch} and TensorFlow is 850~MB in size \cite{tensorflow2015-whitepaper}.

\paragraph{ONNX Runtime.} We therefore used a smaller model interpreter library called ONNX Runtime \cite{bai2019onnx}, which is mere 14~MB in size, leaving a lot of space for the model. Prior to executing the model by the ONNX Runtime library, it needs to be converted to the ONNX format. This can be done using off-the-shelf tools, for instance the Hugging Face \texttt{transformers} library \cite{wolf2020transformers} is shipped with a simple out-of-the-box script to convert BERT models to ONNX.

\paragraph{TensorFlow Lite.} It is also possible to use the TensorFlow Lite interpreter library \cite{tensorflow2015-whitepaper}, which is 6~MB in size. However, we only used ONNX in our deployments as we had problems converting more complex BERT models to TensorFlow Lite format.

\subsection{Serverless deployment}
After training the models and converting them into the ONNX format, we deployed them to different serverless environments.

    \begin{table}
      \begin{adjustbox}{width=1\columnwidth}

      \begin{tabular}{l|SSS|SSS}
        \toprule
        \multirow{2}{*}{} &
          \multicolumn{3}{c}{\textbf{AWS} } &
          \multicolumn{3}{c}{\textbf{GCP} } \\
          & {q50} & {q95} & {q99} & {q50} & {q95} & {q99} \\
          \midrule
        Sentiment TinyBERT & 6.63 & 19.20 & 24.77 & 10.47 & 100.71 & 110.31 \\
        Sentiment MobileBERT & 64.67 & 89.00 & 105.84 & 27.58 & 125.04 & 176.46 \\
        \midrule
        STinyBERT & 5.71 & 13.03 & 21.24 & 10.93 & 101.32 & 111.80 \\
        SMobileBERT & 50.08 & 80.14 & 102.65 & 58.88 & 175.14 & 213.56 \\
        \bottomrule
      \end{tabular}
     \end{adjustbox}
    \caption{Performance comparison between the Amazon Web Services (AWS) and Google Cloud Platform (GCP) serverless environments. Numbers denote execution time in miliseconds with 1GB of RAM allocated for the deployed function. q50, q95 and q99 denote the 0.5, 0.95 and 0.99 quantiles, respectively.}
    \label{tab:aws-gcp-comparison}
    \end{table}

\section{Deployment evaluation}

We measured the performance of deployed models in scenarios with various amounts of allocated memory by making them predict on more than 5000 real-world examples. Before recording measurements we let the deployed model evaluate a small subsample of data in order to keep the infrastructure in a ''warm'' state. This was done in order to estimate the real-life inference time, i.e. to avoid biasing the inference results by initialization time of the service itself.

From the results described in Table~\ref{tab:aws-gcp-comparison} we can see that using both the AWS and GCP platforms, we can easily reach the 0.99 quantile of execution time on the order of 100~ms for both tasks and models. Figure~\ref{fig:all-duration} also lets us observe that the execution time in AWS Lambda decreases with increasing RAM. This is expected, as both AWS Lambda and GCP Cloud Functions automatically allocate more vCPU with more RAM.

The serverless deployments are also cost-effective. The total costs of 1M predictions, taking 100~ms each and using 1~GB of RAM, are around \$2 on both AWS and GCP, whereas the cheapest AWS EC2 virtual machine with 1~GB of RAM costs \$8 per month.

% Tuto by to asi chcelo otestovat nejake modely predtrenovane na GLUE a potom testovat
% - average inference time (per size)
% - cost
% - size
\section{Conclusion}

We present a novel approach of deploying domain-specific BERT-style models in a serverless environment. To fit the models within its limits, we use knowledge distillation and fine-tune them on domain-specific datasets. Our experiments show that using this process we are able to produce much smaller models at the expense of a minor decrease in their performance. The evaluation of the deployment of these models shows that it can reach latency levels appropriate for production environments, while being  cost-effective. 

Although there certainly exist platforms and deployments that can handle much higher load (often times with smaller operational cost \cite{zhang2019mark}), the presented solution requires minimal infrastructure effort, making the team that trained these models completely self-sufficient. This makes it ideal for smaller-scale deployments, which can be used to validate the model's value. The smaller, distilled models created in the process can then be used in more scalable solutions, should the cost or throughput prove inadequate during test deployments.

\bibliography{anthology,naacl2021}
\bibliographystyle{acl_natbib}

\end{document}